# Semantic Segmentation for Urban-Scene Images

**Shorya Sharma**
Indian Institute of Technology Bhubaneswar
ss118@iitbbs.ac.in

## Abstract

Urban-scene Image segmentation is an important and trending topic in computer vision with wide use cases like autonomous driving [1]. Starting with the breakthrough work of Long et al. [2] that introduces Fully Convolutional Networks (FCNs), the development of novel architectures and practical uses of neural networks in semantic segmentation has been expedited in the recent 5 years. Aside from seeking solutions in general model design for information shrinkage due to pooling, urban-scene image itself has intrinsic features like positional patterns [3]. Our project seeks an advanced and integrated solution that specifically targets urban-scene image semantic segmentation among the most novel approaches in the current field. We re-implement the cutting edge model DeepLabv3+ [4] with ResNet-101 [5] backbone as our strong baseline model. Based upon DeepLabv3+, we incorporate HANet [3] to account for the vertical spatial priors in urban-scene image tasks. To boost up model efficiency and performance, we further explore the Atrous Spatial Pooling (ASP) layer in DeepLabv3+ and infuse a computational-efficient variation called "Waterfall" Atrous Spatial Pooling (WASP) [6] architecture in our model. We find that our two-step integrated model improves the mean Intersection-Over-Union (mIoU) score gradually from the baseline model. In particular, HANet successfully identifies height-driven patterns and improves per-class IoU of common class labels in urban scenario like fence and bus. We also demonstrate the improvement of model efficiency with help of WASP in terms of computational times during training and parameter reduction from the original ASPP module.

## 1  Introduction

### 1.1  Background

Semantic image segmentation, the task of labeling each pixel of an image with a corresponding class of what is being represented, has always been a challenging and crucial task in the field of computer vision [7]. **Urban-scene** image segmentation is a particular type that falls into this topic. It has been widely developed in recent years, which expedites applications like autonomous driving vehicles. Take self-driving as an example: the images and videos captured by car-mounted cameras can generally form large scale datasets applicable for deep neural network training. Therefore, advanced deep learning techniques play a significant role in improving segmentation performance for the overall scene background and the individual objects moving in front of the cars.

Starting with the seminal work of Long et al. [2] that introduces Fully Convolutional Networks (FCNs) into semantic segmentation, the development of novel architectures and practical uses of neural networks in semantic segmentation has been expedited in the recent 5 years. Advanced techniques such as skip-connections in encoder-decoder architectures [8] and Atrous Convolution

[9] are further introduced to the FCN-based architecture to resolve multi-scale object and resolution reduction problems. The fruitful variations on model design achieve successful results in diverse semantic segmentation benchmarks [10, 11] including urban-scene datasets.

On the other hand, the urban-scene image is a specific type of image in semantic image segmentation that has intrinsic features regarding positional patterns and geometry knowledge. For example, since the urban-scene images used in autonomous driving usually are captured by the camera positioned at the front of the car, data points are mostly road-driving pictures with spatial positioning bias. In horizontally segmented sections, roads are usually centered, with side-walk and trees at the picture's left and right-hand sides. The spatial prior applies to the vertical position as well: the sky is usually located at the top section, while cars are usually captured at the lower part of the image. With the rapid development of self-driving algorithms, various researches [3] has been conducted recently to account for such information and has proven to contributes significantly to the urban-scene scenario.

### 1.2 Problem Statement

Currently, there are a large amount of model architectures that perform well on general semantic segmentation tasks. Besides, researches done on the distinct nature of urban-scene images yield the possibility of incorporating intrinsic image structural information to these general semantic segmentation models. However, there has not been a thorough and optimal model that infuses the researches on both sides to ensure high performance while maintaining computational efficiency. In this project, we would like to incorporate multiple helpful prior knowledge that applies to urban-scene images. We aim to deploy integrated and advanced deep learning algorithms that target specifically for **urban-scene** image semantic segmentation that searches for a balance between model performance and computational efficiency.

## 2 Literature Review

### 2.1 Advancement in model architecture

The innovations in Convolutional Neural Networks (CNNs) by the authors of [5, 12] form the core of image classification and serve as the structural backbone for state-of-the-art methods in semantic segmentation. However, an important challenge with incorporating CNN layers in segmentation is the significant reduction of resolution caused by pooling. FCN [2] overcame the problem by replacing the final fully-connected layers with deconvolutional stages. It resolves the resolution issues with upsampling strategies across deconvolution layers, increasing the feature map size back to the original image dimensions. The contributions of FCN [2] motivated research in semantic segmentation that led to a variety of different approaches that are visually illustrated in Figure 1.

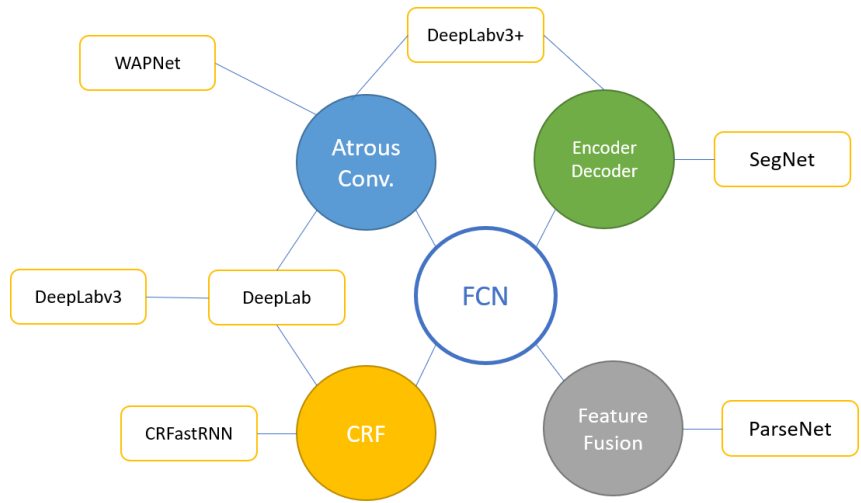

Figure 1: Research Overview in Semantic Segmentation

### 2.1.1 Atrous Convolution

The most popular technique shared among semantic segmentation architectures is the use of dilated or Atrous Convolutions. An early work by Yu et al. [13] highlighted the uses of dilation. Atrous convolutions were further explored by the authors of [14, 15]. Atrous Convolutions' main objectives are to increase the size of the receptive fields in the network, avoid downsampling, and make a generalizable framework for segmentation tasks that can account for the multi-scale object problem. Atrous Convolutions alter the convolutional filters by the insertion of "holes" or zero values in the filter, resulting in the increased size of the receptive field, resembling a hybrid of convolution and pooling layers.

### 2.1.2 Spatial Pyramid Pooling(SPP)

Existing deep convolutional neural networks (CNNs) require a fixed-size input image. This requirement is "artificial" and may reduce the recognition accuracy for the images or sub-images of an arbitrary size/scale. As a result, He et al. [16] equip the networks with another pooling strategy, "spatial pyramid pooling", to eliminate the above requirement. The intuition behind the model design is that SPP perform some information "aggregation" at a deeper stage of the network hierarchy (between convolutional layers and fully-connected layers) to avoid the need for cropping or warping at the beginning.

### 2.1.3 DeepLab

To leverage the generalization power of SPP, Chen, L. et al. [14] has figured out a DeepLab system with Atrous Spatial Pyramid Pooling(ASPP). The special configuration of ASPP assembles dilated convolutions in four parallel branches with different rates. The resulting feature maps are combined by fast bilinear interpolation with an additional factor of eight to recover the feature maps in the original resolution. The main advantages of DeepLab system are improvement on speed, accuracy as well as simplicity.

The application of Atrous Convolution followed the ASPP approach in [14] was later extended in to the cascade approach, that is, the use of several Atrous Convolutions in sequence with rates increasing through its flux. This approach, named Deeplabv3 [9], allows the architecture to perform deeper analysis and increment its performance. Contributions in [9] included module realization in a cascade fashion, investigation of different multi-grid configurations for dilation in the cascade of convolutions, training with different output stride scales for the Atrous Convolutions, and techniques to improve the results when testing and fine-tuning for segmentation challenges.

## 2.2 Specification on Urban-scene Image's Nature

As mentioned before, urban scene image has its intrinsic nature that can be explored and added in the model architecture specifically targets the urban scenario and in terms helps improve the algorithms for autonomous driving. One problem in the urban-scene images is that objects tends to have different scales with small objects like walking people crowded at two sides of the images, and large cars is captured by the car-mounted cameras and is considered as a large object scale. FoveaNet [17] localizes a "fovea region", where the small scale objects are crowded, and performs scale normalization to address heterogeneous object scales. By connecting multiple Atrous Convolutional layers [14, 15] to address large-scale changes of the objects, DenseASPP [18] is proposed to adopts densely connected ASPP. By exploiting which classes appear frequently at a particular position in an image, spatial priors can be learnt to improve urban-scene segmentation tasks. Choi et al. propsoed a Height-driven Attention Net (HANet) [3] that uses the vertical information, and this is the context prior that we would like to implement in our model architecture. Also, a class-balanced self-training with spatial priors [19] generates pseudo-labels for unlabeled target data in the field of domain adaption to aid the development in unsupervised learning.

# 3 Contribution

To seek the optimal solution specifically for **urban-scene** images semantic segmentation tasks, we researched the fruitful amount of novel methodology in the most recent years in the field. We cross-compare with different *state-of-the-art* model performances and choose DeepLabv3+ as our strong baseline model [4], which is one of the distinguished and efficient models for semantic segmentation. After that, we exploit the positional pattern of urban-scene images using HANet [3] and an advanced design of Atrous Spacial Pooling layer in model architecture called WASP [6] to improve model performance. Eventually, we propose an integrated model to achieve our project objectives in (1) targeting urban scene nature and (2) maintaining model simplicity and efficiency. Our final model architecture is illustrated in the figure below.

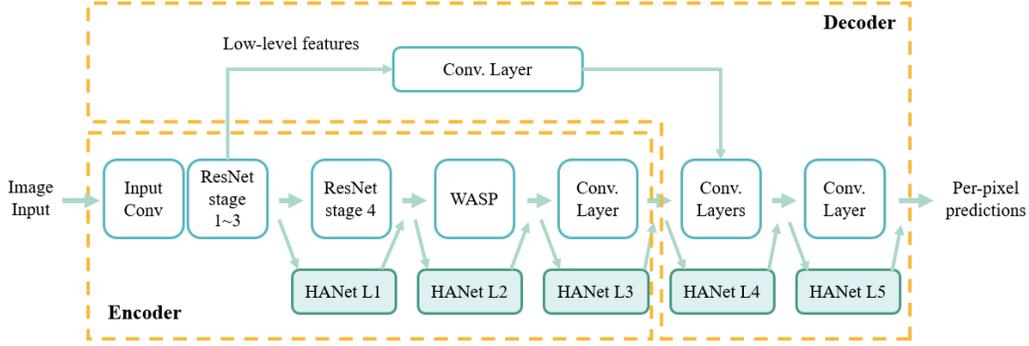

Figure 2: Our Integrated Model Architecture

Our project contributes in the following points:

1. Re-implementation of our strong baseline model DeepLabv3+.

2. Variation 1: DeepLabv3+ with HANet to see incorporate height-driven positional pattern in the model design. HANet is a flexible add-on module to build on various model architecture. We re-implement the idea on our chosen baseline model.

3. Variation 2: DeepLabv3+ with HANet and WASP: change the ASPP layer with WASP layer in Deeplabv3+ to reduce computational complexity. WASP is a novel idea as an advanced alternative of Atrous Spatial Pooling (ASPP) layer. We propose the idea of replacing WASP with the original ASPP layer without harming our model performance.

**3.1 DeepLabv3+: Encoder-Decoder with Atrous Separable Convolution**

DeepLabv3+ [4] is an advanced network system developed derived from the FCN framework. It highlights an encoder-decoder network on top of the atrous convolution to help with faster computation and obtaining more precise and sharper object edges.

In semantic segmentation task, the Encoder-Decoder system is usually consisted of: (1) an encoder module that gradually reduces the feature maps and captures higher semantic information, and (2) a decoder module that gradually recovers the spatial information.

DeepLabv3+ has demonstrated its ability to produce sharper object boundaries with their proposed Atrous Spatial Pyramid Pooling (ASPP) layers in the encoder module and send the information to Decoder to recover the edge information. More specifically, the output stride (ratio of input image spatial resolution to the final output resolution) is usually 32 for the image classification tasks. However, in the semantic segmentation task, we need denser pixel-wise information of a feature map to produce good object boundaries. So DeepLabv3+ uses an output stride of 16 or 8 to form encoder features. Then in the proposed decoder, the encoder features are first bilinearly upsampled by a factor of 4 and then concatenated with the corresponding low-level features to ensure enough spatial information is preserved.

With DeepLabv3 as the powerful encoder and a simple yet effective decoder, DeepLabv3+ is able to combine the advantages of both models and achieve a prominent improvement in the precision and speed of various computer vision objectives, including object detection and semantic segmentation.

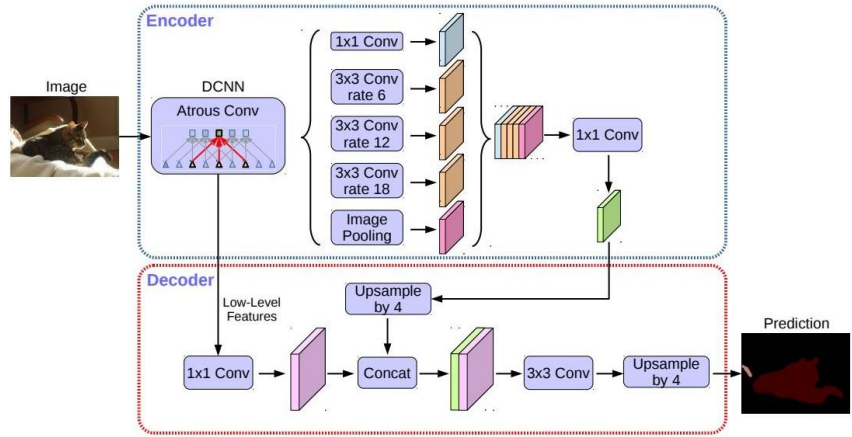

Figure 3: Baseline Model - DeepLabv3+ Pipeline Illustration Retrieved From Paper [4]

## 3.2 HANet: Height-driven Attention Networks

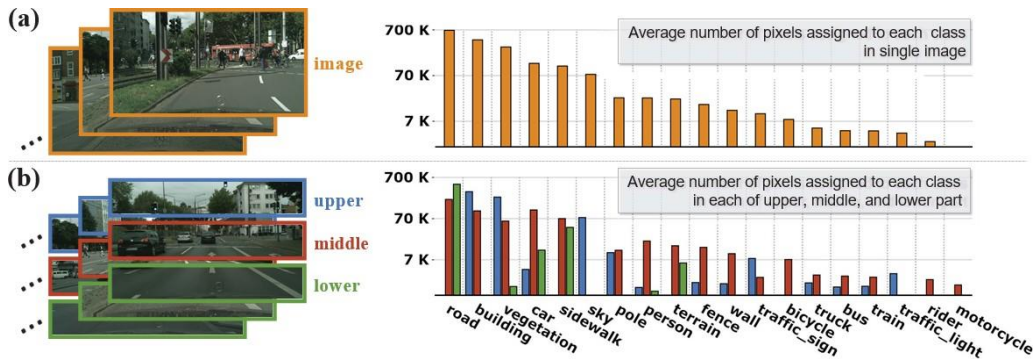

Figure 4: Pixel-wise Class Distributions of Cityscapes Dataset [20] Retrieved from Paper [3]

Urban-scene images have their special perspective geometry and positional patterns. Since the urban-scene images are captured by the cameras mounted on the front side of a car, the urban-scene datasets consist only of road-driving pictures. This leads to the possibility of incorporating common structural priors depending on a spatial position, markedly in a vertical position. To verify this characteristic, Fig.4 presents the class distribution of an urban-scene dataset across vertical positions. From the picture, we can see that the class distribution has significant dependency on a vertical position. The lower part of an image is mainly composed of road, while the middle part contains various kinds of relatively small objects. In the upper part, buildings, vegetation, and sky are principal objects.

Inspired with such observation, we seek a way of incorporating vertical spatial information into the network in recognizing different objects in urban scene setting. We found that Choi et al. [3] propose a novel architecture Height-Driven Attention Netowork (HANet) as a general add-on module to semantic segmentation for urban-scene images. Given an input feature map, HANet extracts "height-wise contextual information", which represents the context of each horizontally divided part, and then predicts which features or classes are more important than the others within each horizontal part from the height-wise contextual information.

HANet generates per-channel scaling factors for each individual row from its height-wise contextual information as its architecture illustrated in Fig.5.

Let $X_l$ and $X_h$ denote the lower and higher-level feature maps in semantic segmentation networks, where C is the number of channels, $H$ and $W$ are the spatial dimensions of the input tensor, height and width, respectively. Given the lower-level feature map $X_A$, $F_{HANet}$ generates a channel-wise attention map $A$ made up of height-wise per-channel scaling factors and fitted to the channel and height dimensions of the higher-level feature map $X_h$. This is done in a series of steps: width-wise pooling (Fig.5(a)), interpolation for coarse attention (Fig.5(b,d)), and computing height-driven attention map (Fig.5(c)). Moreover, adding positional encoding is included in the process (Fig.5(e)).

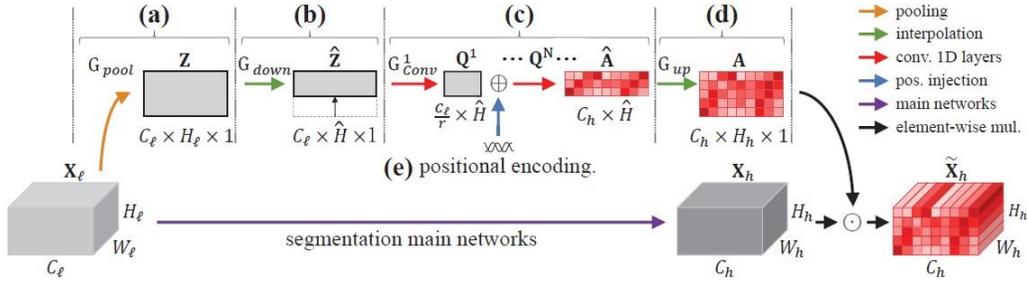

Figure 5: Architecture of HANet Retrieved from Paper [3]

Specifically, the way that HANet is incorporated into the baseline model is by adding the sinusoidal positional encoding, which is defined as

$$PE_{(p,2i)} = sin(p/100^{2i/C})$$
$$PE_{(p,2i+1)} = cos(p/100^{2i/C})$$

Where p denotes the vertical position index in the entire image ranging from zero to $\hat{H}$ - 1 of coarse attention, and i is the dimension. The number of the vertical position is set to $\hat{H}$ as the number of rows in coarse attention. The dimension of the positional encoding is same as the channel dimension C of the intermediate feature map $Q_i$. The new representation $\tilde{Q}$ incorporating positional encoding is formulated as

$$\tilde{Q} = PE \oplus Q$$

Where $\oplus$ is an element-wise sum. This is the way that positional information is encoded into the feature map like demonstrated in Fig.5(e).

With the advantage of HANet, we decided to add it on top of our baseline model DeepLabv3+. By adding HANet to our baseline model(DeepLabv3+), we postulate that the location information can improve the model result.

### 3.3 "Waterfall" Atrous Spatial Pooling Architecture(WASP): Efficient way to maintain Field-of-View

Our second project objective is to optimize our model design in terms of model architecture design. DeepLabv3+ is an outstanding model that can incorporate with different backbones like ShuffleNet, ResNet-38 and ResNet-101. Although DeepLabv3+ outperforms other state-of-the-art models, the model itself with deep backbone is very complicated and suffers from long computational time in training. Through studying the architecture of our model and also reviewing some related work, we find out that there is a possibility to optimize the time efficiency of Atrous Spatial Pyramid Pooling (ASPP) layer in DeepLabv3+.

Inspired by the cascaded architectures and multiscale approaches, Artacho and Savakis [6] further propose a new "Waterfall" configuration for the ASPP layers to overcome the complexity and memory issues with ASPP, called Waterfall Atrous Spatial Pooling (WASP). Figure 6 below shows a brief comparison between the ASPP module and WASP module. The ASPP module employed a parallel architecture for the branches, with no parameters shared and connected in between; whereas the WASP architecture uses a waterfall-like way to sequentially forward the intermediate results to the next branch. By using this structure, larger Field-of-View can be fed in the network and less parameters will be required. According to the experiments conducted by the author [6], they successfully gained 20.69% reduction of the parameters with a 2% boosting of the model performance (mIoU) using the WASPnet they built upon WASP.

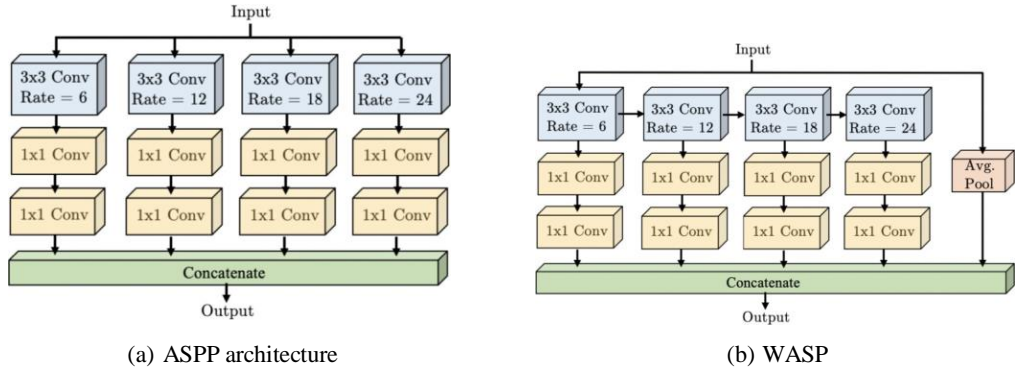

(a) ASPP architecture    (b) WASP

Figure 6: Comparison of the Architecture of ASPP and WASP Modules

Therefore, we considered to replace the ASPP module in the original DeepLabv3 model with the state-of-art WASP module, trying to experiment the compatibility and robustness of the WASP module on different backbones, as well as its effects on the model performance.

## 4 Experiment

In this section, we first describe the implementation details of our two-step integrated models. We ran our experiment on the Cityscapes [20] dataset. For all the quantitative experiments, we measure the segmentation performance in terms of mean Intersection over Union (mIoU) metric. We would also elaborate on the selection of our strong baseline model.

### 4.1 Experimental Setting

Based on our two variations in model design, the experiments aims to answer the following two questions from our hypothesis:

1. Does HANet successfully identity height-driven information and in terms improve model performance in urban-scene setting?
2. Does WASP reduce the computational efficiency in training duration while showing possibility in boosting or maintaining model performance?

#### 4.1.1 Experiment with Two-step Variation Models

We first validate our baseline model DeepLabv3+. We build our baseline model code by adopting open-source resources from the author's GitHub [1].

Then, we run our experiment to add HANet on top of DeepLabv3+ to test hypothesis (1). The HANet is a generalizable add-on module that can be fit into various model architecture. We adopt the HANet module from original GitHub [2] and infused in our codebase.

Lastly, we conduct another experiment to change ASPP module with WASP in order to test hypothesis (2). The WASP module is inspired by GitHub [3] and incorporated into our model.

### 4.1.2 Training Process Overview

**Data Preprocessing & Data Augmentation** Our first challenge during experiment is to increase the diversity of data available for the training and to avoid overfitting. As a result, We perform data augmentation on the Cityscapes dataset by cropping images into 512*1024, random horizontally flipping, random scaling, Gaussian blur and color jittering to increase the diversity of data available for the training and avoid overfitting.

**Customization in Loss** At initial development stage, we used standard cross entropy loss. However, it does not give us a very good performance result due to the imbalanced issue in dataset. In urban scene datset, class like roads are highly dominated, which bias the training procedure using standard cross entropy. Inspired by PSPNet [21], we use a cross entropy loss with customized weight for different class label to address the issue and boost model performance. We also adopt an auxiliary Cross-Entropy Loss in the intermediate feature map and class uniform sampling to handle this problem. To adopt the auxiliary loss, we have added additional convolutional layers to our backbone ResNet stage 3 as an auxiliary branch. The loss for this auxiliary branch has a weight of 0.4.

**Optimization Settings** For each controlled experiment, we use SGD optimizer with parameter listed in table 1 below. The learning rate scheduling follows the polynomial learning rate policy [22]. The other hyper-parameter inside Deeplabv3+, HANet and WASP is suggested by original papers to ensure the optimal results for training. When we first train our model, we use a g4dn.4xlarge instance. However, CUDA went out of memory because of the limitation of GPU capacity. After searching on Google, we found that expanding GPU devices can solve the problem. Finally, we figure out that a AWS g4dn.12xlarge (4 GPUs) instance works best in our case.

| Network | Backbone | Model | output stride | learning rate | mome-ntum | weight decay | num epoch | machine |
|---|---|---|---|---|---|---|---|---|
| DeepLabv3+ | ResNet101 | Baseline | 16 | 0.01 | 0.9 | 0.0005 | 240 | g4dn.12xlarge |
| DeepLabv3+ | ResNet101 | HANet | 16 | 0.01 | 0.9 | 0.001 | 230 | g4dn.12xlarge |
| DeepLabv3+ | ResNet101 | HANet+WASP | 16 | 0.01 | 0.9 | 0.0005 | 230 | g4dn.12xlarge |

Table 1: Experimental Setting for Three Experiment Variations.

## 4.2 Dataset

### 4.2.1 Cityscapes Dataset

**Cityscapes** The dataset we will be primarily using is Cityscapes [20], a diverse large-scale dataset designed for urban scene semantic segmentation. It is derived from video sequences recorded in the streets of 50 cities. It contains 5K images with high-quality pixel-level annotations and 20K images with coarse annotations (Figure 7).

We use the fine annotation set in our experiments. The fine annotation set with 5k data points is then split into a training set (2,975 images), a validation set (500 images), and a test set (1525 test images), and 19 semantic labels are defined.

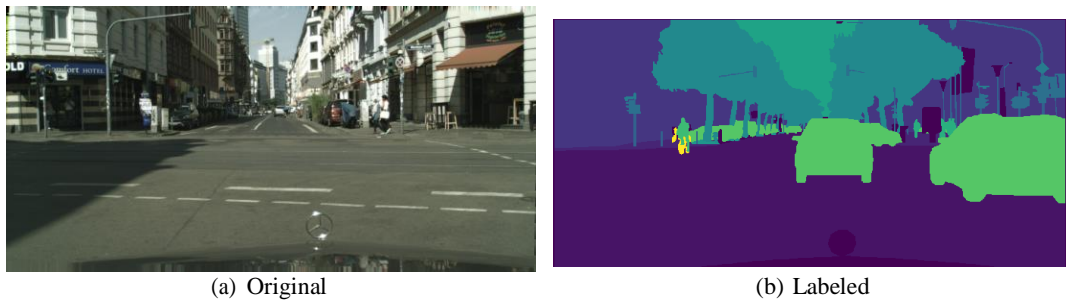

(a) Original     (b) Labeled

Figure 7: Cityscapes Dataset Demo

### 4.2.2 Data Pre-processing & Data Augmentation

The Cityscapes dataset we use in the experiments contains 5K high-quality images. Considering the complexity of objects in urban scene scenarios and the possibility of overfitting, data augmentations will be very important in our case to increase the diversity of data available for the training. Inspired by the methodologies used in NVIDIA segmentation [23], we adopted a combination of data augmentation techniques in our model, such as random horizontally flipping, random scaling, Gaussian blur and color jittering.

**Random Horizontally Flipping** We conducted a 0.5 random horizontally left to right flipping on the dataset to maintain the invariance of directions.

**Random Scaling** A random scaling for the size of the images was also conducted so that different scales of each object can be presented to the model to increase the invariance of images with different resolutions.

**Gaussian Blur** Gaussian Blur will blur an image using a Gaussian distribution so that it can reduce the noises and negligible details in images. We used it as an important technique in our case to smooth the images and intensify the image structures with different object scales.

**Color Jittering** To simulate and generate urban scene images under different lighting environments, we employed color jittering to randomly adjust the hue, brightness, contrast and saturation of the color channels.

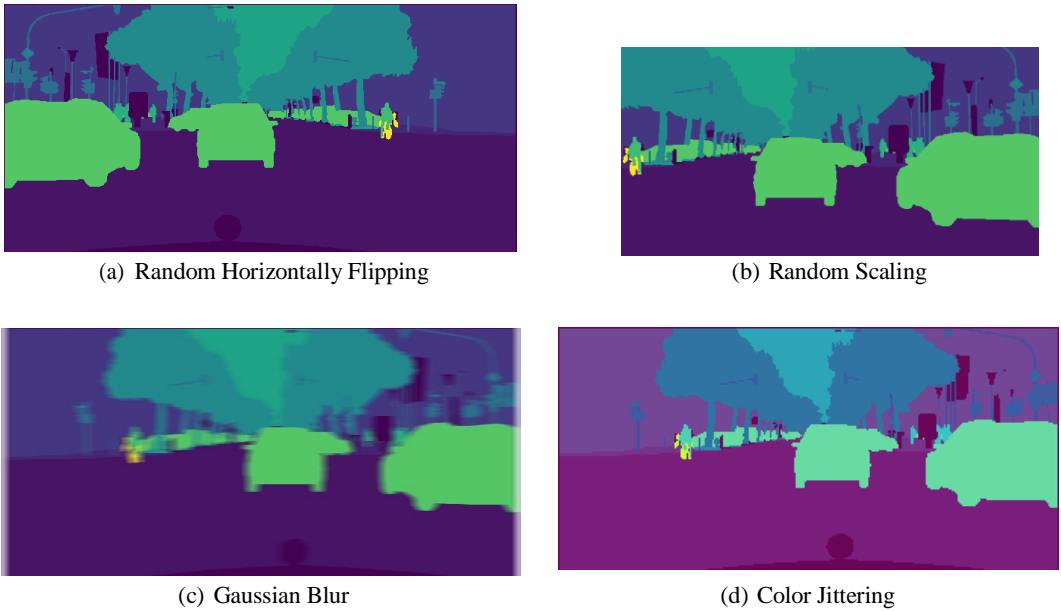

(a) Random Horizontally Flipping  (b) Random Scaling

(c) Gaussian Blur  (d) Color Jittering

Figure 8: Data Augmentation Demo

## 4.3 Evaluation Metrics

We based our comparison of performance with other methods on Intersection over Union (IOU), as it is the most important and more widely used metric for semantic segmentation. Specifically, we monitor both per-class Intersection-Over-Union (pIoU) and Mean Intersection-Over-Union (mIoU) as evaluation metrics, to see how the model is performed for different label classes.

### 4.3.1 Intersection-Over-Union(IoU)

The Intersection-Over-Union(**IoU**), as known as Jaccard Index, is calculated by the number of overlapping pixels between the predicted segmentation and the ground truth divided by the number of union pixels of predicted segmentation and the ground truth. Figure 9(a) provides a visualized calculation of **IoU** scores. For multi-class segmentation in our project, we can calculate **per-class IoU** and also **mean IoU** (mIoU), which is taking the average of **per-class IoU**.

A **IoU** score is a range between 0 and 1, with 0 meaning totally wrong prediction and 1 meaning perfectly correct prediction. As **IoU** appreciated corrected labeled portion by accounting for overlap, it is a less biased measurement in general cases. One possible limitation is at **IoU** does not necessarily tell you how accurate the segmentation boundaries are[24].

### 4.3.2 Exploration of other evaluation metrics

IoU is better than other common metrics like pixel accuracy, which measures the proportion of correctly labeled pixels. Pixel accuracy in general is a highly limited metric that yields biased and uninformative impressions for imbalanced dataset and is not considered in our project.

Dice Coefficient(F1 score) is also a relatively good metric to consider in the case of unbalanced dataset, which is exactly our case. It is calculated by twice the area of overlap divided by the total number of pixels in both images (See Figure 9(b) for visualized illustration). The reason that we choose IoU over Dice Coefficient as our evaluation metric is that (1) IoU can also accounts for imbalanced dataset issue and have similar purpose as Dice Coefficient (2) most of the state-of-the-art model uses mIoU score to evaluate model and we want to be consistent in the evaluation metrics we used throughout the project.

Besides, since our project aims to see the influence of positional patterns in the model architecture, we especially calculate per-class IoU besides mean IoU to more precisely monitor and analyze our performance results.

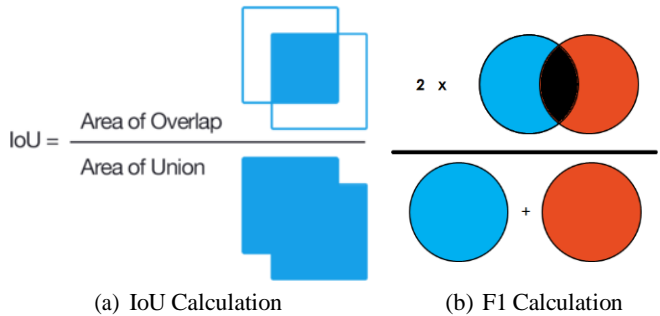

(a) IoU Calculation      (b) F1 Calculation

Figure 9: IoU Calculation vs F1 Calculation. Retrieved from Wikipedia.

## 4.4 Baseline Model: DeepLabv3+

By comparing and evaluating the benefits and limitations of current novel and efficient state-of-the-art models we have research upon, we selected **DeepLabv3+** [4] as our baseline model. The baseline model performance results that we ran is similar to the published results, DeepLabv3+ outperforms among the existing state-of-the-art models [2, 4].

| Network | Backbone | wall | road | fence | bus | sign | bike | person | light | mIoU |
|---|---|---|---|---|---|---|---|---|---|---|
| FCN-8s | ResNet50 | 41.4 | 97.9 | 56.5 | 75.1 | 79.4 | 77.5 | 82.3 | 71.4 | 72.4 |
| FCN-8s | ResNet101 | 45.6 | 98.2 | 62.1 | 70.9 | 80.2 | 78.9 | 83.5 | 72.8 | 75.2 |
| DeepLabv3+ | ResNet50 | 51.2 | 98.2 | 57.9 | 77.3 | 77.2 | 74.1 | 85.9 | 73.5 | 76.8 |
| DeepLabv3+ | ResNet101 | 54.8 | 98.4 | 58.8 | 77.9 | 78.5 | 75.3 | 86.0 | 74.2 | 77.8 |

Table 2: Comparison of mIoU and Per-class IoU in Percentage on Baseline Models with Different Backbones on **Cityscapes** Dataset

Table 2 below shows the baseline models with different backbones in the evaluation metrics of mIoU in percentage. The baseline models can be based on different backbones, such as ResNet-38, ResNet-50, ResNet-101. We choose DeepLabv3+ with ResNet-101 as backbones, given its optimal model performances in mIoU in different backbones [4].

We found out that DeepLabv3+ outperforms among the state-of-the-art models in mIoU and per-class IoU scores. Therefore, we view DeepLabv3+ as an outstanding achievement in urban-scene image semantic segmentation and plan to use it as a stronger baseline model in future experiments. As mentioned before, modification like positional prior add-on module [3] can be generalized on top of various models. Therefore we can flexibly incorporate and experiment with the variation on top of DeepLabv3+ to see whether model variations have add-up influences on model performance results.

## 5 Model Results

### 5.1 Analysis of IoU for different objects and mIoU

The comparison of mIoU and per-class IoU result between DeepLabv3+, HANet+DeepLabv3+ and WASP+HANet+DeepLabv3+ models is as follow:

| Network | Backbone | Model | wall | road | fence | bus | sign | bike | person | light | mIoU |
|---|---|---|---|---|---|---|---|---|---|---|---|
| DeepLabv3+ | ResNet101 | Baseline | 54.8 | 98.4 | 58.8 | 77.9 | 78.5 | 75.3 | 86.0 | 74.2 | 77.8 |
| DeepLabv3+ | ResNet101 | HANet | 60.8 | 98.6 | 65.7 | 91.6 | 80.8 | 79.4 | 83.6 | 72.8 | 80.9 |
| DeepLabv3+ | ResNet101 | HANet+WASP | 63.4 | 98.4 | 63.9 | 92.4 | 80.0 | 79.1 | 83.9 | 72.3 | 81.0 |

Table 3: Comparison of mIoU and Per-class IoU in Percentage with DeepLabv3+, HANet+DeepLabv3+ and WASP+HANet+DeepLabv3+ Models on **Cityscapes** Dataset

From the Table 3, we can see that both HANet+DeepLabv3+ and WASP+HANet+DeepLabv3+ outperform our baseline DeepLabv3+ overall. Especially, they gain a great improvement for objects like wall, fence and bus. This is because HANet makes use of height-driven positional information and improves the detection of smaller objects that lies in specific horizontal subsections. However, for some taller classes like light, person and pole, the performance of HANet gets worse. One possible reason is that HANet splits the images into subsections and the structure of taller objects across multiple sections might be disrupted. In the next section, we further analyze the results through visualization.

### 5.2 Analysis of Visualization Results

Figure 10 below shows a demo of our modeling result, including the original image, color-masked prediction and composed version. Generally, our two variants of the model perform better than the baseline in urban-scene images in semantic segmentation, and this can be specified in three different aspects.

**Objects with smaller size** One key improvement of adding HANet on the baseline is that it can greatly improve the segmentation performances on smaller objects. For example, in Fig.10 (b), the traffic sign on the right side is not very clearly detected. However, for our two variants, the traffic sign is detected with more precise boundaries. This supports the hypothesis of HANet that adding the height-driven spatial prior of objects will improve the model performance on specific classes [3]. In the high-level sense, HANet split the height of images into several sections and train a height-driven attention map for different height sections. Small objects normally appear in the lower and upper horizontal sections, therefore we can expect better results.

**Objects under poor lighting or with blurry edges** Another improvement of our two variants on the baseline is that we achieve better performance for objects under poor lighting or with blurry edges. For example, we could see in Fig.10 (b) that the two pedestrians in the left shadow are not properly segmented; whereas in Fig.10 (e) and Fig.10 (h), they are clearly and fully detected. It appears to us that the positional encoding in HANet helps in generalizing and clearing the vertically split boundaries. Besides, performing data augmentation techniques like Color Jittering and Gaussian

Blur also helps the model to recognize the dark and blur boundaries and enhances the overall model performance.

**Objects with taller size** We also find out that HANet doesn't perform well on taller objects, like lights and poles (as shown in Fig.10 (e) and Fig.10 (h)). As mentioned in previous analysis, HANet splits the image vertically into high, middle and low section to account for objects' distribution in different sections. However, taller objects like road lights normally locates across multiple sections and the intrinsic structure of taller objects is being disrupted in the model design of HANet. Therefore, the effect of positional information in HANet is quite limited to those objects.

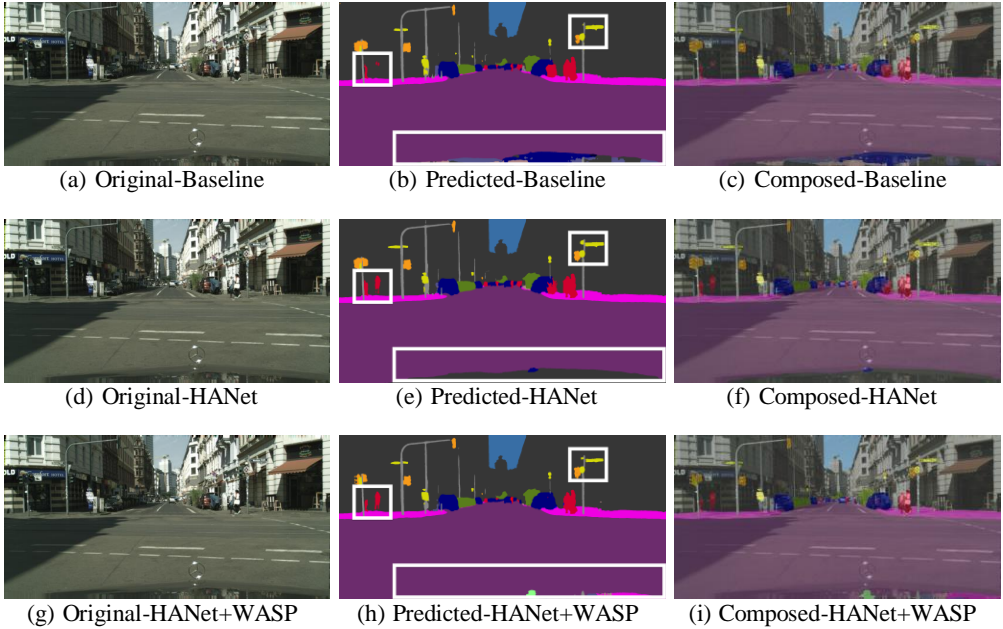

(a) Original-Baseline  (b) Predicted-Baseline  (c) Composed-Baseline

(d) Original-HANet  (e) Predicted-HANet  (f) Composed-HANet

(g) Original-HANet+WASP  (h) Predicted-HANet+WASP  (i) Composed-HANet+WASP

Figure 10: Baseline and HANet Model Results

### 5.3 Analysis of Time and Memory

We further evaluate the time and memory required for each model in the experiments. As shown in the table below, we can see that the average minutes per epoch for our baseline and the first variant are pretty similar, which aligns with the conclusion in HANet [3] that adding a HANet module on the Deepladv3+ backbone won't harm the computation cost for training but can generate better IoU results [3].

| Network | Backbone | Output Stride(os) | Model | mIoU | Min per Epoch |
|---|---|---|---|---|---|
| DeepLabv3+ | ResNet101 | 16 | Baseline | 77.8 | 4.5 |
| DeepLabv3+ | ResNet101 | 16 | HANet | 80.9 | 4.5 |
| DeepLabv3+ | ResNet101 | 16 | HANet+WASP | 81.0 | 4 |

Table 2: Comparison of mIoU and Per-class IoU in Percentage with DeepLabv3+ and HANet+DeepLabv3+ Models on **Cityscapes** Dataset

Another finding we can gain from the table is that replacing ASPP blocks with WASP generates similar results, but requires less memory and time. The mIoU for AASP + HANet and WASP + HANet are quite close to each other. However, We successfully gain a 12.5% reduction of the average minutes per epoch with the WASP + HANet variant. Thus, we think WASP can work as a good proxy when processing capability is limited during training with traditional Deeplabv3+.

# 6 Conclusion & Future Work

## 6.1 Conclusion

In this project, we develop an integrated model based on the most novel and efficient discoveries in the field to improve the performance in urban scene image segmentation. We incorporate HANet on DeepLabv3+ to exploit the vertical priors existing in urban scene images. We further improve our architecture's computation efficiency by replacing the Atrous Spatial Pyramid Pooling (ASPP) module in the backbone with Waterfall Atrous Spatial Pooling (WASP) module. Last but not the least, advanced data augmentation techniques are also adopted to increase the diversity of our dataset and avoid overfitting.

By conducting the experiment, we first validate hypothesis (1) that vertical position based HANet as an add-on module effectively boosts model performance by monitoring per-class IoU and mean IoU improvement. We observe that added HANet increases the mean IoU to 80.9% to from 77.8% in baseline model. Additionally, height driven positional patterns is proved to be captured as we show the improvement of per-class IoU in smaller objects that lie within the same horizontal sections and objects under poor lighting or with blurry edges. The hypothesis (2) is also validated as we gain a 12.5% reduction of the average minutes per epoch through replacing the ASPP module with WASP, without hurting the performance. Thus, it proves that WASP can work as a good proxy for ASPP under limited processing capability.

## 6.2 Future Works

- **Experiment of new architecture on more models**: In our current experiment, we apply WASP and HANet to DeepLab v3+ and the new architecture improved the performance a lot. However, we are not sure about whether applying WASP + HANet architecture is a universal method to improve the performance. Thus, we need explore WASP + HANet architecture on more models such as PSPNet, CGNet and FCN-8s for next step. Also, we can explore the architect on the same model with different backbones like ResNet 50 and Xception.

- **Improvement on specific class identification**: Although our model got awesome performance overall, the identification of taller object like person and light gets worse. Therefore, how to correctly identify taller objects is our next challenge. On one hand, we can make use of horizontal context to improve the model. In our initial trial, we attempted Reality Oriented Adaptation to learning the horizontal context, but the result is not as expected. In the future, we will try more models to combine horizontal context. On the other hand, we can also attempt to combine tall object identification model with current model.

- **Parameter reduction**: One application of semantic segmentation for urban-scene images is autonomous driving, which requires rapid inference of the picture. Currently, it takes about 0.3 second for our model to infer one urban-scene image.To apply our model to autonomous driving, we need to improve the speed of inference. One possible direction of the improvement is parameter reduction, which can reduce both computational time and memory. To implement parameter reduction, we can try some new architectures or optimization method.

- **Coarsely-Annotated Data and Auto-labelling**: The Cityscapes Dataset has another 20K coarsely-annotated image data. Due to the time, memory and computation power constraint of this project, we we have not taken into consideration in our project. By adding coarse annotated set, there would be obvious amount of increase of our data point, so we would possibly expect an increase model performance from the current standpoint. The coarse annotation set itself has limitation that requires further pre-processing before feeding into the training procedure. As it is only coarsely annotated, use them cannot fully make use of the all label information, and requires extra work in considering refining the coarse annotation. We do see examples of other excellent research conducted on the Cityscapes dataset and incorporate the 20k coarsely annotated set, with an increase of around 1-2% in mIoU scores. Also, we see a helpful research conducted by NVIDIA Segmentation [23] that they proposed a hard-threshold based auto-labelling strategy which leverages unlabeled images and boosts IOU. This is something that in the future works could be incorporated and considered in our project in order to making full use of the information from coarse annotation images.

# References

[1] Andreas Ess, Tobias Mueller, Helmut Grabner, and Luc Van Gool. Segmentation-based urban traffic scene understanding. In Andrea Cavallaro, Simon Prince, and Daniel C. Alexander, editors, *British Machine Vision Conference, BMVC 2009, London, UK, September 7-10, 2009. Proceedings*, pages 1–11. British Machine Vision Association, 2009.

[2] J. Long, E. Shelhamer, and T. Darrell. Fully convolutional networks for semantic segmentation. In *2015 IEEE Conference on Computer Vision and Pattern Recognition (CVPR)*, pages 3431–3440, 2015.

[3] Sungha Choi, Joanne T. Kim, and Jaegul Choo. Cars can't fly up in the sky: Improving urban-scene segmentation via height-driven attention networks. In *Proceedings of the IEEE/CVF Conference on Computer Vision and Pattern Recognition (CVPR)*, June 2020.

[4] Liang-Chieh Chen, Y. Zhu, G. Papandreou, Florian Schroff, and H. Adam. Encoder-decoder with atrous separable convolution for semantic image segmentation. In *ECCV*, 2018.

[5] K. He, X. Zhang, S. Ren, and J. Sun. Deep residual learning for image recognition. In *2016 IEEE Conference on Computer Vision and Pattern Recognition (CVPR)*, pages 770–778, 2016.

[6] Bruno Artacho and Andreas E. Savakis. Waterfall atrous spatial pooling architecture for efficient semantic segmentation. *CoRR*, abs/1912.03183, 2019.

[7] Jeremy Jordan. An overview of semantic image segmentation., Nov 2020.

[8] Vijay Badrinarayanan, Alex Kendall, and Roberto Cipolla. Segnet: A deep convolutional encoder-decoder architecture for image segmentation. *CoRR*, abs/1511.00561, 2015.

[9] Liang-Chieh Chen, George Papandreou, Florian Schroff, and Hartwig Adam. Rethinking atrous convolution for semantic image segmentation. *CoRR*, abs/1706.05587, 2017.

[10] M. Everingham, L. Gool, C. K. Williams, J. Winn, and Andrew Zisserman. The pascal visual object classes (voc) challenge. *International Journal of Computer Vision*, 88:303–338, 2009.

[11] Gerhard Neuhold, Tobias Ollmann, Samuel Rota Bulò, and Peter Kontschieder. The mapillary vistas dataset for semantic understanding of street scenes. In *International Conference on Computer Vision (ICCV)*, 2017.

[12] K. Simonyan and Andrew Zisserman. Very deep convolutional networks for large-scale image recognition. *CoRR*, abs/1409.1556, 2015.

[13] Fisher Yu and Vladlen Koltun. Multi-Scale Context Aggregation by Dilated Convolutions. *arXiv e-prints*, page arXiv:1511.07122, November 2015.

[14] L. Chen, G. Papandreou, I. Kokkinos, K. Murphy, and A. L. Yuille. Deeplab: Semantic image segmentation with deep convolutional nets, atrous convolution, and fully connected crfs. *IEEE Transactions on Pattern Analysis and Machine Intelligence*, 40(4):834–848, 2018.

[15] Liang-Chieh Chen, George Papandreou, Florian Schroff, and Hartwig Adam. Rethinking atrous convolution for semantic image segmentation, 2017.

[16] Kaiming He, Xiangyu Zhang, Shaoqing Ren, and Jian Sun. Spatial pyramid pooling in deep convolutional networks for visual recognition. *CoRR*, abs/1406.4729, 2014.

[17] Xin Li, Zequn Jie, Wei Wang, Changsong Liu, Jimei Yang, Xiaohui Shen, Zhe Lin, Qiang Chen, Shuicheng Yan, and Jiashi Feng. Foveanet: Perspective-aware urban scene parsing. In *IEEE International Conference on Computer Vision, ICCV 2017, Venice, Italy, October 22-29, 2017*, pages 784–792. IEEE Computer Society, 2017.

[18] Maoke Yang, Kun Yu, Chi Zhang, Zhiwei Li, and Kuiyuan Yang. Denseaspp for semantic segmentation in street scenes. In *Proceedings of the IEEE Conference on Computer Vision and Pattern Recognition (CVPR)*, June 2018.

[19] Yang Zou, Zhiding Yu, B.V.K. Vijaya Kumar, and Jinsong Wang. Unsupervised domain adaptation for semantic segmentation via class-balanced self-training. In *Proceedings of the European Conference on Computer Vision (ECCV)*, September 2018.

[20] Marius Cordts, Mohamed Omran, Sebastian Ramos, Timo Rehfeld, Markus Enzweiler, Rodrigo Benenson, Uwe Franke, Stefan Roth, and Bernt Schiele. The cityscapes dataset for semantic urban scene understanding. In *Proc. of the IEEE Conference on Computer Vision and Pattern Recognition (CVPR)*, 2016.

[21] Hengshuang Zhao, Jianping Shi, Xiaojuan Qi, Xiaogang Wang, and Jiaya Jia. Pyramid scene parsing network. In *CVPR*, 2017.

[22] Wei Liu, Andrew Rabinovich, and Alexander C. Berg. Parsenet: Looking wider to see better. *CoRR*, 2015.

[23] Andrew Tao, Karan Sapra, and Bryan Catanzaro. Hierarchical Multi-Scale Attention for Semantic Segmentation. *arXiv e-prints*, page arXiv:2005.10821, May 2020.

[24] Ekin Tiu. Metrics to evaluate your semantic segmentation model, Oct 2020.